# Reduced Deep Convolutional Activation Features (R-DeCAF) in Histopathology Images to Improve the Classification Performance for Breast Cancer Diagnosis

**Bahareh Morovati, Reza Lashgari, Mojtaba Hajihasani and Hasti Shabani \***

Institute of Medical Science and Technology, Shahid Beheshti University, Tehran, Iran
\* Correspondence: ha_shabani@sbu.ac.ir

**Abstract:** Breast cancer is the second most common cancer among women worldwide. Diagnosis of breast cancer by the pathologists is a time-consuming procedure and subjective. Computer aided diagnosis frameworks are utilized to relieve pathologist workload by classifying the data automatically, in which deep convolutional neural networks (CNNs) are effective solutions. The features extracted from activation layer of pre-trained CNNs are called deep convolutional activation features (DeCAF). In this paper, we have analyzed that all DeCAF features are not necessarily led to a higher accuracy in the classification task and dimension reduction plays an important role. Therefore, different dimension reduction methods are applied to achieve an effective combination of features by capturing the essence of DeCAF features. To this purpose, we have proposed reduced deep convolutional activation features (R-DeCAF). In this framework, pre-trained CNNs such as AlexNet, VGG-16 and VGG-19 are utilized in transfer learning mode as feature extractors. DeCAF features are extracted from the first fully connected layer of the mentioned CNNs and support vector machine has been used for binary classification. Among linear and nonlinear dimensionality reduction algorithms, linear approaches such as principal component analysis (PCA) represent a better combination among deep features and lead to a higher accuracy in the classification task using small number of features considering specific amount of cumulative explained variance (CEV) of features. The proposed method is validated using experimental BreakHis dataset. Comprehensive results show improvement in the classification accuracy up to 4.3% with less computational time. Best achieved accuracy is 91.13% for 400× data with feature vector size (FVS) of 23 and CEV equals to 0.15 using pre-trained AlexNet as feature extractor and PCA as feature reduction algorithm.

## 1. Introduction

Breast cancer (BC) is one of the leading causes of mortality in the world almost observed in women, but it can occur in men, too. Diagnosis of BC ordinarily comprises of an initial detection by palpation and regular check-ups by ultrasound imaging or mammography and diagnosis of possible malignant tissue growth is tested by breast tissue biopsy [1]. According to the world health organization (WHO), BC is affecting large number of women health [2]. Recent studies predict around 27 million new cases of BC by 2030 [3]. Early detection of BC is essential for appropriate treatment and decreasing the mortality rate. However, BC diagnosis may not be accurate enough as pathologist could only apply visual inspection of samples under microscopes [4, 5]. According these challenges, computer aided diagnosis and automatic classification using convolutional neural networks (CNNs) for image classification are an active research area to make a precise diagnosis with less probability of misdiagnosis and fast detection process.

Current state-of-the-art investigations on BC detection confirm that CNNs are more reliable and faster than the conventional hand-crafted features in the classification task [6]. However, estimated time to train CNNs might be longer and it needs expertise to design such networks [6-8]. An applicable solution reported in the literature is referred as deep convolutional activation feature (DeCAF) also known as deep features [6,

8, 9]. These approaches reuse pre-trained CNNs to extract deep features and apply them to a classifier for final decision.

The hand-crafted features in BC histopathological dataset (737 images) has been studied by Filipczuk *et al*. [10] using circular Hough transform to segment the cell nuclei by circles. Their best result reached 98.51% accuracy utilizing k-nearest neighbor (KNN) as a classifier [10]. However, the region of interest in the virtual slides is not selected automatically and it is a time-consuming process. Additionally, the method cannot guarantee a global optimum and elliptical segmentation requires more accurate model which is computationally more demanding. In another work by Sharma and Mehra, hand-crafted features like color, shape and texture extracted from BreakHis dataset and fed them to the conventional classifiers such as support vector machine (SVM) and random forest (RF). They reported RF with 1000 number of trees could achieve 90.33% accuracy for 40× data [11]. In addition, they have compared the hand-crafted features with deep ones. The accuracy obtained for the classification of the deep features using VGG-16 network for 40× data is 93.97%. They have reported that the performance of the hand-crafted features is not satisfactory since it requires deep knowledge about the morphology of cancerous cells and deep features are a preferred alternative. Alhindi *et al*. compared the local binary patterns (LBP), the histogram of oriented gradients (HOG) as the hand-crafted features with deep features using the pre-trained VGG-19 for KIMIA Path960 dataset [12]. The highest accuracy is 90.52% for LBP features and SVM classifier. It is worth mentioning that feature vector size (FVS) of LBP is equal to 1182 and almost twice of the one of the extracted deep features.

To address deep features in histopathological images, Spanhole *et al*. extracted DeCAF from different fully connected (FC) layers of pre-trained AlexNet with logistic regression classifier to diagnose BC using BreakHis dataset [6]. The obtained results show that transfer learning is a viable alternative with 84.6% accuracy for 40× data. Then, Deniz *et al*. developed a framework to take advantage of two pre-trained CNNs for binary classification of BreakHis dataset. They have combined DeCAF features from AlexNet and VGG-16 followed by SVM classifier and reached 84.87% accuracy [8]. In [13], Kumar *et al*. proposed a variant of VGG-16, wherein all FC layers were removed and evaluated by different classifiers for CMT and BreakHis datasets. The best reported accuracy is 97.01% for 200× data from BreakHis dataset, in which the FVS is 1472.

To overcome the lack of training dataset, dividing the histopathological image into non-overlapping or random patches and providing them as the input to the pre-trained CNNs for feature extraction has been studied [1, 6, 14]. However, extracting some patches can lead to uncertainty of the classification [15]. To improve the accuracy of the classification, some approaches focused on training CNNs from scratch or fine-tuning the pre-trained CNNs [15-19]. Some of these approaches have been reached to the higher performance while experiencing a time-consuming procedure and arranging hyperparameters precisely. In some cases, training the model or fine-tuning all the layers may not achieve a better performance compared to transfer learning technique [7, 16, 17]. Additionally, transfer learning hits the spot either encountering the lack of training dataset to train a deep model or adding a few training data to re-train the whole model [13, 14, 16].

Dimension reduction or feature selection of deep features has attracted the attention of researchers recently. Alinsaif *et al*. applied Infinite Latent Feature Selection (ILFS) method to select top ranked features from pre-trained CNNs such as ResNet and DenseNet w/wo fine-tuning. The accuracy of binary classification for BreakHis dataset with SVM classifier is reported 97.96% where FVS is 1300 [16]. Moreover, Gupta *et al*. proposed extreme gradient boosting (XGboost) to reduce the number of features extracted from ResNet and used information theoretic measure (ITS) to select the optimal number of layers. The accuracy is reported 97.07±1.18% for 40× data where FVS is 500. Although, the accuracy decreased with fewer number of features [20].

In this study, dimensionality reduction is the main scope to investigate the influence on capturing informative features with a smaller number of features. We have analyzed that all the deep features are not necessarily led to a higher accuracy in the classification task and dimension reduction plays an important role. We have proposed R-DeCAF features to capture the essence of the data with low computational time. To achieve such a milestone, the weight of the pre-trained CNNs (AlexNet, VGG-16 and VGG-19) will be kept in freeze mode and deep features extracted from the first FC layer, in which the size of the feature vector is high, i.e., FVS = 4096. In order to reduce the size of the extracted deep features, different linear and nonlinear dimension reduction methods such as PCA, singular value decomposition (SVD), linear discriminant analysis (LDA), kernel PCA (kPCA) and t-Distributed Stochastic Neighbor Embedding (t-SNE) have been evaluated to generate R-DeCAF features. The comprehensive comparison of DeCAF and R-DeCAF features (which are reduced by linear methods) classified by SVM with RBF kernel on BreakHis dataset shows keeping less than 120 features not only improves the classification accuracy but also decreases the computational time.

This paper is organized as follows: Section 2 describes the histopathological dataset; Section 3 illustrates the problem formulation and provides details of the proposed model to extract R-DeCAF. In section 4, the experimental case studies and comparative analysis of the obtained results are discussed. Finally, section 5 represents the conclusion and the future research.

## 2. Histopathological database

The BreakHis database [21] developed in a laboratory (Pathological Anatomy and Cytopathology, Parana, Brazil) and it is a publicly accessible histopathologic BC dataset which used in this work. This selected dataset, includes microscopic histopathology images of BC, consists of 7,909 images of BC tissue taken from 82 patients which is available in 40×, 100×, 200× and 400× magnification factors. This dataset includes 2480 benign and 5429 malignant samples with the color image size of 700×460. In addition, benign and malignant tumors are divided into subgroups. Samples of this dataset are collected by surgical open biopsy (SOB) method and stained by Hematoxylin and Eosin method. Each image filename, includes stored information about the image such as biopsy procedure method, magnification factor, type of cancer and its subtypes, and patient identification. In Fig. 1, sample images of benign and malignant tumor from this dataset at different magnification factors are shown.

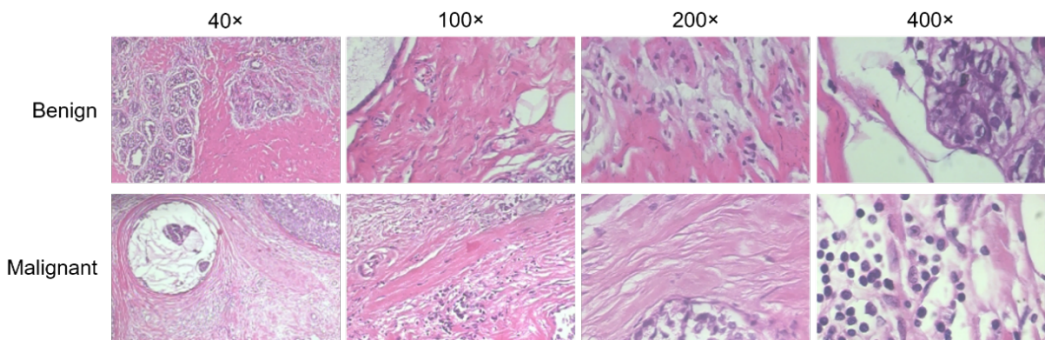

**Figure 1.** Sample images from BreakHis database in different magnification factors. First row belongs to the same slide of the benign tumor and second row belongs to the same slide of the malignant tumor.

## 3. Methodology

This study involves deep feature extraction from BreakHis histopathological dataset and we have shown that all the features extracted from the pre-trained CNNs cannot be effective in classifying the data. Therefore, reducing the FVS of the extracted deep features to keep informative features and remove unnecessary ones which cause misleading or do not play an important role in the classification, is the primary goal of this study. Different contemporary pre-trained CNN models involving deep feature extraction from BreakHis dataset are considered. Pre-trained CNNs have been trained by ImageNet dataset, which contains more than 14 million natural images with 1000 categories.

The advantage of transfer learning technique is to avoid a time-consuming procedure for training a CNN from scratch [6, 13, 16]. In other words, training and fine-tuning a CNN requires a huge amount of data or a medium size of data, respectively. However, transfer learning involves different structures and it does not need a huge amount of data. Actually, transfer learning is a method to transfer the knowledge learned in one domain to a defined task for the purpose of classification or feature extraction. Thus, the goal is transferring the knowledge from natural images to BC histopathological images and simplify the diagnosis process. As pre-trained CNNs are trained on a large dataset with significant number of classes and samples, it can provide large networks with effective combination of features to classify the data [6, 13].

Deep features are extracted automatically from the first FC layer of AlexNet, VGG-16 and VGG-19 networks in transfer learning mode with considering the weights in freeze mode. Thus, freezing the weights trained based on ImageNet dataset makes the model prepared to use all defined pre-trained weights. The reason of considering the first FC layer i.e. FC6 of AlexNet, VGG-16 and VGG-19 is that FC6 layer provides features more informative for an accurate classification [6, 22].

*3.1. Proposed framework for R-DeCAF*

We reduce FVS of DeCAF features which is fixed to 4096 by applying appropriate dimension reduction algorithms to generate R-DeCAF features, in which FVS is less than 120. According to the redundancy or misleading features in a data [23], and based the analysis that we have done all DeCAF features are not required in the classification task. Moreover, a high dimensional feature vector can dramatically impact the performance of machine learning algorithms to fit on data and generally this can referred to as the "curse of dimensionality" [24]. Therefore, we have proposed R-DeCAF features to capture the essence of the data with low computational time by analyzing different dimensionality reduction techniques. The architecture of the proposed framework is illustrated in Fig. 2.

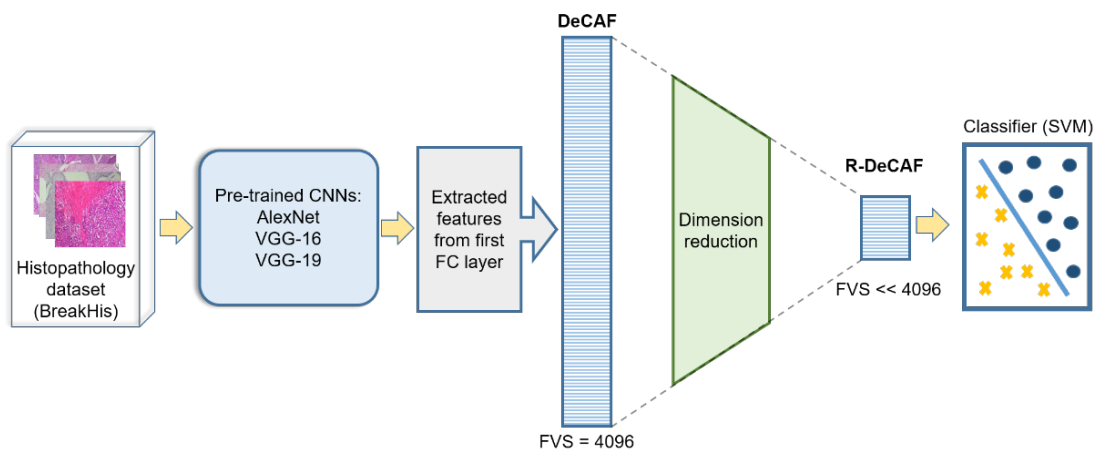

**Figure 2.** The diagram of the proposed framework.

*3.2. Pre-trained CNNs*

All three defined pre-trained CNNs which are AlexNet, VGG-16 and VGG-19 are studied as the basic of our framework to extract deep features. AlexNet is known to be the primary profound CNN model presented by Krizhevsky *et al.* [25]. This network contains five convolutional layers and three FC layers where the number of neurons in the last layer is based on the number of classes of the data. The number of neurons in first and second FC layers are 4096. The VGG-16 and VGG-19 CNNs with more layers are proposed by Simonyan *et al.* in 2014 [26]. In these two CNNs, small filters of 3×3 are used for all the layers in order to capture fine details in the images and control the number of parameters. VGG-19 has 19 weight layers and VGG-16 has 16 weight layers [26]. It should be mentioned that all the input images are resized to 224×224 for the sake of convenience with CNN models in pytorch library in this work.

*3.3. Feature Reduction Algorithms*

This study analyzes different dimension reduction methods on DeCAF features categorized in two groups; linear and nonlinear. The former includes PCA, SVD, and LDA, where the later contains kPCA and t-SNE [23, 27]. PCA method is a linear and unsupervised algorithm, in which new features can be produced by calculating a linear transformation. Eigenvectors and eigenvalues can be computed from the covariance matrix of the data to determine the *principal components* (PC) of the data. PCA keeps the maximum information of the data in the first PC and continues in descending order because principal directions and corresponding PCs are considered as the directions of the maximum data variance [23, 24]. SVD is another linear dimension reduction method which is appropriate for sparse data. SVD of a matrix is a factorization of the main matrix into three matrices. In this method, the largest singular values are picked, where the eigenvalues and eigenvectors are in descending order same as PCA method. Hence, the input matrix will rebuild in low dimension [23]. LDA is another linear and supervised dimension reduction method which focuses on two critical terms called "scatter between class" and "scatter within class". The main aim is to maximizing the "scatter between class" or separability of classes. Therefore, LDA can pick components which separates the data classes in the best way. It should be mentioned that the number of components/features in a reduced dimension can be equal or smaller than the number of classes-1 [23, 28].

In the group of nonlinear dimension reduction methods, kPCA is one of the popular unsupervised techniques. When the PCA method does not work well and the structure of the data is nonlinear, kPCA method may perform better. In kPCA, the dimension of the original data can be reduced in a high dimensional space with the advantage of "kernel trick". In high dimensional space decision boundary becomes linear. In this method the eigenvalues and eigenvectors of the kernel matrix are calculated based on the reduced dimension set of eigenvectors selected in descending order. The product of the original matrix and eigenvectors is calculated to rebuild the new reduced data [24, 27]. The nonlinear and unsupervised t-SNE method is known as a common technique for data exploration and visualization. In this method, data is mapped to a low dimension, such as 2 or 3 dimensions. t-SNE converts the high dimensional Euclidean distances between pairwise data points $x_i, x_j$ into conditional probability $p_{j|i}$ which shows the similarity of the pairwise data points and a similar conditional probability in low dimensional counterparts $y_i, y_j$ of the high dimensional data points $x_i, x_j$ defined by $q_{j|i}$. The conditional probabilities $p_{j|i}$ and $q_{j|i}$ will be equal if the data points $y_i, y_j$ model the similarity between the data points in a high dimensional space [23, 29].

*3.4. Cumulative Explained Variance (CEV) and the size of the features*

To reduce the size of the features, we have used eigenvalues and the corresponding cumulative explained variance (CEV). Actually, determining the optimal number of PCs is a challenging and a critical key in order to get an efficient performance and CEV is a way to solve this challenge. CEV is the accumulation of variances to show the summation of variances of the new features i.e., PCs as the percentage of this accumulated variance by the PC numbers [23]. Figure 3 displays CEV of DeCAF features extracted from the first FC layer of pre-trained AlexNet, VGG-16 and VGG-19 which is related to the whole magnification data of BreakHis dataset. As it can be seen, approximately with more than 2560 PCs, the CEV has changed insignificantly. In other words, the first 2560 PCs contains 100% of cumulative variances where the first 512 PCs covers 67% of variance of data.

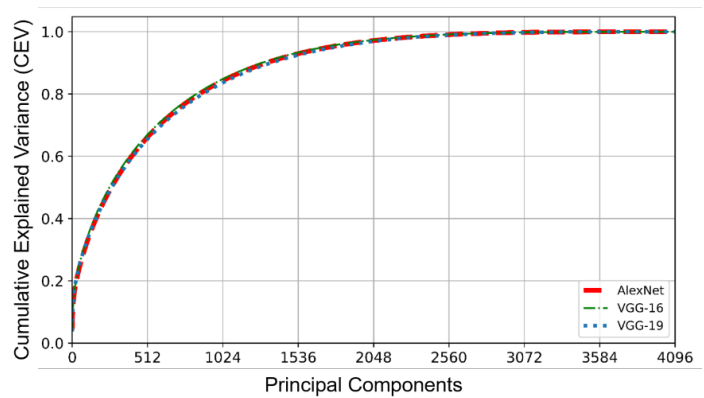

**Figure 3.** CEV of DeCAF features using pre-trained CNNs for the whole magnification data of BreakHis dataset.

It can be concluded that almost half of the transformed features i.e., PCs do not have important role in the classification as a rule of thumb where Zhong *et al.* [9] took advantage of this simple rule. The main reason is because of high correlation among extracted deep features. To investigate more in details, we have considered a full range of CEV from 5% to 100%, in which 100% means we have used all PCs obtained from DeCAF features to examine the classification accuracy. Here, the average accuracy is obtained for three pretrained CNNs in 10 different splits of feature vectors into train and test datasets. Figure 4 shows these results for the whole magnification data (7909 images) of BreakHis dataset.

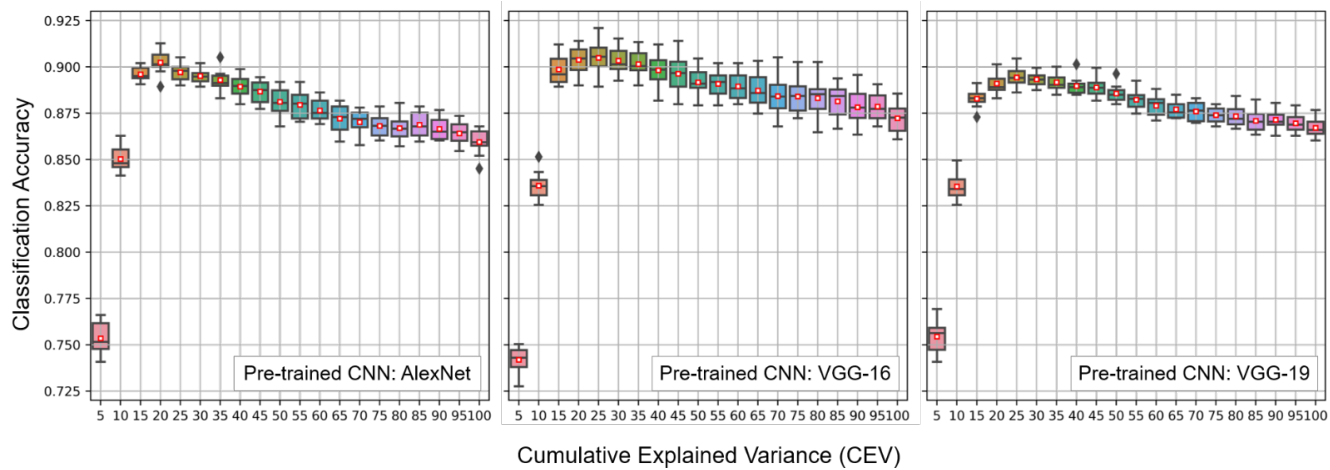

**Figure 4.** The average classification accuracy vs Cumulative Explained Variance (CEV) for the whole magnification data of BreakHis dataset.

First, these results confirm that using CEV less than 100% but more than 15% not only keep the same performance but also causes improvement in classifying deep features. Second, the results of Fig. 4 provide more information that we need to feed the classifier with more effective and proper features rather than large number of features. Therefore, a better accuracy can be achieved with a smaller number of features and the content of features plays a crucial role in the classification task. We have shown by these results that considering large number of features could not necessarily lead to a higher performance and all DeCAF features extracted from pre-trained AlexNet, VGG-16, and VGG-19 are not compelling and informative in the classification. Figure 4, discloses keeping only 20% to 25% of CEV makes more improvement compared to 50% of CEV. These also reduced FVS significantly from 4096 to 63, 103, and 93 for the pre-trained AlexNet, VGG-16, and VGG-19, respectively. More details of this investigation are presented in Table. 1.

*3.5. Classifier*

Here, the SVM algorithm has been selected for the classification as it has the ability to handle the high dimensional data and nonlinear classification by using a kernel trick [13]. This technique is used to evaluate the performance of DeCAF and R-DeCAF features in classification task to predict a sample is benign or malignant. The trained SVM with RBF kernel is considered as the common kernels-based on a Grid search among different kernels with the SVM parameter C=5. The defined dataset is divided into a training set (80%) and test set (20%). The split method is used and the results are reported by taking an average of 10 different splits. Comprehensive results are provided by pytorch and Scikitlearn libraries to validate the proposed method. Since most of machine learning algorithms are sensitive to data scaling, in this manner we apply Standard Scalar of Scikitlearn library to scale the feature vectors that are extracted.

## 4. Results and Discussion

The classification accuracy of DeCAF and R-DeCAF features are summarized in Table. 1. R-DeCAF features obtained by three linear dimension reduction algorithms, i.e., PCA, SVD and LDA. First, we have investigated the classification performance using DeCAF features by three mentioned CNNs in more details. As you can see in third column of Table.1, the accuracy of VGG-16 and VGG-19 outperform AlexNet for 40×, 100×, and 200× data of BreakHis dataset but underperform for 400×. The reason can be found in both the number of layers and 3×3 filters in VGG-16 and VGG-19 networks which can extract more details form images. However, for 400× data such details from VGG-16 and VGG-19 networks are not necessary as the magnification is higher and the images provide such details. Therefore, AlexNet is a better choice for high magnification data. It is worth mentioning that the accuracy considering the whole magnification is better for VGG-16 and VGG-19 networks as expected. In addition, the lowest accuracy is observed for a 40× data. This might be because of the region of interest of 40× data as includes a higher complexity compared with other magnification factors and carries more information which makes the accurate data classification more difficult. The magnification factor effect on the classification accuracy depending on complexity level of BC histopathological images.

Second, the classification performance using R-DeCAF features based on different dimension reduction algorithms has been explored further. The results from applying PCA and SVD to generate R-DeCAF features are provided almost the same improvement up to 4.3% compared to DeCAF features. For example, the observed accuracy considering pre-trained AlexNet for whole magnification data are 85.95%, 90.24% and 90.18 for DeCAF features, R-DeCAF features using PCA, and R-DeCAF features using SVD, respectively. Again, we can say that AlexNet is still a better choice for high magnification data even for R-DeCAF features. The results obtained by LDA depict the classification accuracy has been decreased and this method is not able to capture the essence of data caused by removing informative features. The main reason explaining the low accuracy by LDA is that the number of features of original dataset is ignored and the obtained dimension (FVS) will be less than the number of classes subtract one. Therefore, we will have only one feature for binary classification using LDA technique [23].

**Table 1.** The classification accuracy (%) of DeCAF and R-DeCAF features using different linear dimension reduction methods (Best results are bolded).

| Framework | Magnification | DeCAF FVS = 4096 | R-DeCAF (reduced by linear methods) | | | | |
|---|---|---|---|---|---|---|---|
| | | | FVS (CEV) | PCA | FVS (CEV) | SVD | LDA |
| AlexNet (FC6), SVM | 40× | 84.38±1.6 | 67 (0.25) | **88.04±2.1** | 67 (0.25) | 88.02±2.1 | 77.62±1.8 |
| | 100× | 86.16±1.1 | 45 (0.20) | 89.54±1.0 | 45 (0.20) | **89.66±0.9** | 79.87±2.3 |
| | 200× | 87.69±1.4 | 97 (0.30) | **90.60±1.4** | 135 (0.35) | 90.50±1.3 | 83.87±1.4 |
| | 400× | 87.91±1.5 | 23 (0.15) | 91.13±1.4 | 23 (0.15) | **91.15±1.5** | 82.58±1.5 |
| | Whole mag. | 85.95±0.8 | 63 (0.20) | **90.24±0.6** | 63 (0.20) | 90.18±0.6 | 71.73±1.1 |
| VGG-16 (FC6), SVM | 40× | 86.64±2.2 | 90 (0.30) | **89.82±1.7** | 90 (0.30) | 89.60±1.7 | 82.23±0.9 |
| | 100× | 89.52±1.2 | 58 (0.25) | 91.01±0.7 | 206 (0.45) | **91.01±1.2** | 84.05±1.8 |
| | 200× | 88.71±0.9 | 84 (0.30) | 90.78±1.4 | 56 (0.25) | **90.82±1.1** | 82.68±0.9 |
| | 400× | 85.60±1.9 | 58 (0.25) | **88.30±1.3** | 58 (0.25) | 88.21±1.3 | 81.95±1.9 |
| | Whole mag. | 87.23±0.8 | 103 (0.25) | **90.61±0.9** | 103 (0.25) | 90.35±0.8 | 73.10±1.0 |
| VGG-19 (FC6), SVM | 40× | 85.09±1.8 | 78 (0.30) | **87.34±2.0** | 51 (0.25) | 87.14±1.6 | 79.62±1.9 |
| | 100× | 88.06±1.5 | 118 (0.35) | **90.19±1.3** | 118 (0.35) | 90.05±1.3 | 81.10±1.3 |
| | 200× | 88.51±1.0 | 114 (0.35) | **89.35±1.2** | 114 (0.35) | 89.60±1.6 | 82.80±1.3 |
| | 400× | 86.73±0.8 | 59 (0.25) | 88.60±0.8 | 59 (0.25) | **88.76±0.9** | 82.03±1.9 |
| | Whole mag. | 86.68±0.5 | 93 (0.25) | **89.43±0.5** | 93 (0.25) | 89.37±0.5 | 72.03±0.8 |

To evaluate our proposed method, we have computed the confusion matrix of the classification result for DeCAF and R-DeCAF features. These matrixes are shown in Fig. 5, in which the results obtained by pre-trained AlexNet for 400× data (1820 images, 588 benign and 1232 malignant) of BreakHis dataset. It can be seen that the classification result of R-DeCAF features obtained by PCA outperforms the one of DeCAF features. In more details, correctly predicted benign cases increased from 68% to 82%. This is very impressive result as our proposed method could increase the accuracy in benign class although the number of the data in this class is more limited due to imbalanced BreakHis dataset.

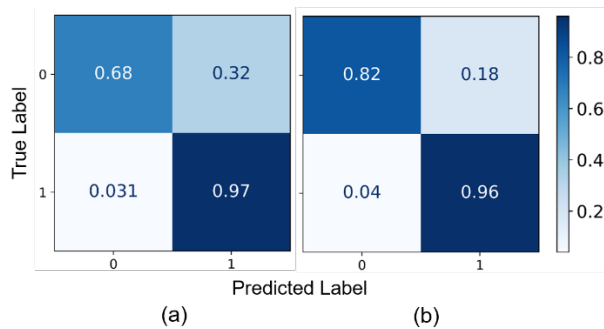

**Figure 5.** Confusion matrix for 400× data form BreakHis dataset to show the classification result of deep features extracted from pre-trained AlexNet. (a). DeCAF features, (b). R-DeCAF features using PCA algorithm (FVS=23, CEV=0.15). 0: Benign, 1: Malignant.

Moreover, we have validated the classification result of DeCAF and R-DeCAF features using other metrics as shown in Table. 2. We have reported R-DeCAF features that reduced by PCA algorithm as we have observed that the accuracy obtained by PCA outperforms other linear dimension reduction algorithms (see Table. 1). Table. 2 shows that the precision and F1 score of R-DeCAF features have been improved compared to the ones of DeCAF features, however recall is different and it has been decreased for R-DeCAF features in some cases.

Since BreakHis dataset is imbalanced in which the number of the samples in the malignant class is almost twice the one in the benign class and this ratio is almost the same for different magnifications, we have addressed this issue by reproducing the results based on two strategies. First, we used weighted SVM and the data is divided into train and test dataset with stratified k-fold (k=10). Second, we forced the data to be balanced by randomly selected malignant samples to be the same as the number of benign samples in each magnification factor and SVM is used. Both results show that the classification metrics in Table. 1 are higher and the effect of imbalance data is not significant as the imbalance ratio of the data is not too high. These investigations also confirm that the balance data could affect recall and improve it however its overall improvement is less compared to Table. 2.

**Table 2.** The classification accuracy, precision, recall and F1 score (%) of DeCAF and R-DeCAF features (reduced by PCA).

| Framework | Magnification | DeCAF (FVS = 4096) | | | | R-DeCAF (reduced by PCA) | | | | |
|---|---|---|---|---|---|---|---|---|---|---|
| | | Accuracy | Precision | Recall | F1 | FVS (CEV) | Accuracy | Precision | Recall | F1 |
| AlexNet (FC6), SVM | 40× | 84.38±1.6 | 85.20±2.1 | 93.42±1.0 | 89.10±1.2 | 67 (0.25) | **88.04±2.1** | **89.61±2.7** | **94.73±1.5** | **91.54±1.6** |
| | 100× | 86.16±1.1 | 85.77±1.5 | **95.95±0.8** | 90.57±0.9 | 45 (0.20) | **89.54±1.0** | **90.05±1.9** | 95.52±1.3 | **92.68±0.7** |
| | 200× | 87.69±1.4 | 87.63±1.5 | 95.58±1.5 | 91.42±0.9 | 97 (0.30) | **90.60±1.4** | **91.11±1.8** | **95.65±1.3** | **93.31±1.0** |
| | 400× | 87.91±1.5 | 87.32±2.1 | **96.02±1.2** | 91.44±1.2 | 23 (0.15) | **91.13±1.4** | **91.31±1.6** | 95.96±0.9 | **93.57±1.1** |
| | Whole mag. | 85.95±0.8 | 86.98±0.9 | 93.35±0.6 | 90.05±0.5 | 63 (0.20) | **90.24±0.6** | **91.24±0.6** | **94.78±0.7** | **92.97±0.5** |
| VGG-16 (FC6), SVM | 40× | 86.64±2.2 | 88.09±2.7 | 93.17±1.9 | 90.53±1.7 | 90 (0.30) | **89.82±1.7** | **90.43±1.9** | **95.28±1.8** | **92.77±1.3** |
| | 100× | 89.52±1.2 | 89.21±1.4 | **96.53±1.2** | 92.72±0.8 | 58 (0.25) | **91.01±0.7** | **92.15±1.1** | 95.11±1.5 | **93.59±0.6** |
| | 200× | 88.71±0.9 | 89.23±1.5 | 95.07±1.6 | 92.04±0.6 | 84 (0.30) | **90.78±1.4** | **91.03±1.8** | **96.04±1.1** | **93.46±1.0** |
| | 400× | 85.60±1.9 | 85.43±2.2 | 95.08±1.4 | 89.98±1.5 | 58 (0.25) | **88.30±1.3** | **88.92±1.8** | **95.64±1.4** | **91.67±1.0** |
| | Whole mag. | 87.23±0.8 | 87.86±0.9 | 94.40±0.6 | 91.01±0.5 | 103 (0.25) | **90.61±0.9** | **91.30±0.8** | **95.18±0.8** | **93.20±0.7** |
| VGG-19 (FC6), SVM | 40× | 85.09±1.8 | 86.00±2.4 | 93.51±1.0 | 89.58±1.3 | 78 (0.30) | **87.34±2.0** | **87.01±2.2** | **95.87±1.3** | **91.21±1.5** |
| | 100× | 88.06±1.5 | 87.33±2.0 | **96.86±0.9** | 91.84±1.1 | 118 (0.35) | **90.19±1.3** | **90.60±1.5** | 95.81±1.0 | **93.13±1.0** |
| | 200× | 88.51±1.0 | 88.33±1.0 | **96.31±0.9** | 92.14±0.6 | 114 (0.35) | **89.35±1.2** | **90.10±1.6** | 95.29±1.2 | 92.61±0.8 |
| | 400× | 86.73±0.8 | 86.59±1.5 | **95.35±1.3** | 90.74±0.6 | 59 (0.25) | **88.60±0.8** | **90.27±0.6** | 93.36±1.4 | **91.78±0.6** |
| | Whole mag. | 86.68±0.5 | 87.39±0.5 | 94.10±0.7 | 90.62±0.5 | 93 (0.25) | **89.43±0.5** | **89.93±0.6** | **95.22±0.5** | **92.50±0.4** |

In addition, this study evaluates the performance of nonlinear dimension reduction methods, including kPCA and t-SNE described in Section 3.3. The classification accuracy based on DeCAF and R-DeCAF features is presented in Table. 3 in addition to FVS. Using kPCA algorithm, a different number of features have been tested. However, the classification accuracy based on R-DeCAF features is not high enough. So, we have considered the same number of features similar to PCA method. To apply t-SNE algorithm, it is highly recommended to first use PCA method before decreasing the dimension to 2 or 4 features by t-SNE. Therefore, we have reduced FVS in the same step as applying only PCA method on the feature vectors. Then, t-SNE is implemented to reduce the number of features to 2. As it is clearly shown, nonlinear dimension reduction methods are not effective to capture informative features from DeCAF features and classification accuracy has decreased. However, linear approaches such as PCA could represent a better combination among deep features and lead to a higher accuracy in the classification task. Nonlinear dimensionality reduction techniques might be sensitive to the curse of dimensionality and this could be the reason of their improper performance in our study. Hence, these methods are not able to guarantee better performance than linear ones, such as PCA [23, 27]. Moreover, we can consider the presence of more complexity in R-DeCAF features obtained by nonlinear dimension reduction methods which lead to lower classification accuracy.

**Table 3.** The classification accuracy (%) of DeCAF and R-DeCAF features using different nonlinear dimension reduction methods (Best results are bolded).

| Framework | Magnification | DeCAF FVS = 4096 | R-DeCAF (reduced by nonlinear methods) | | |
|---|---|---|---|---|---|
| | | | FVS (CEV) | KPCA | PCA+ t-SNE |
| AlexNet (FC6), SVM | 40× | **84.38±1.6** | 67 (0.25) | 68.45±2.6 | 65.94±5.7 |
| | 100× | **86.16±1.1** | 45 (0.20) | 69.33±1.9 | 68.85±1.5 |
| | 200× | **87.69±1.4** | 97 (0.30) | 68.58±1.8 | 65.23±4.8 |
| | 400× | **87.91±1.5** | 23 (0.15) | 67.47±2.4 | 67.03±4.7 |
| | Whole mag. | **85.95±0.8** | 63 (0.20) | 68.15±1.0 | 62.96±6.0 |
| VGG-16 (FC6), SVM | 40× | **86.64±2.2** | 90 (0.30) | 68.77±2.8 | 63.51±11.2 |
| | 100× | **89.52±1.2** | 58 (0.25) | 69.11±1.0 | 68.94±3.2 |
| | 200× | **88.71±0.9** | 84 (0.30) | 68.67±1.5 | 63.57±10.9 |
| | 400× | **85.60±1.9** | 58 (0.25) | 68.21±2.5 | 64.78±6.9 |
| | Whole mag. | **87.23±0.8** | 103 (0.25) | 68.46±0.7 | 66.15±2.7 |
| VGG-19 (FC6), SVM | 40× | **85.09±1.8** | 78 (0.30) | 68.65±2.2 | 55.93±13.20 |
| | 100× | **88.06±1.5** | 118 (0.35) | 69.45±2.0 | 63.86±7.8 |
| | 200× | **88.51±1.0** | 114 (0.35) | 69.95±1.4 | 68.39±3.0 |
| | 400× | **86.73±0.8** | 59 (0.25) | 68.27±1.9 | 66.73±5.4 |
| | Whole mag. | **86.68±0.5** | 93 (0.25) | 68.44±0.9 | 66.96±2.5 |

Based on the analysis that we have done and the results reported in Table 1 and 3, we can conclude that the accuracy of binary classification for BreakHis dataset will be enhanced using R-DeCAF features with linear dimension reduction algorithms like PCA and SVD up to 4.3% in different magnification factors. Less probability of overfitting and noise rejection capability of PCA algorithm and the benefits of sparse data management by SVD algorithm [23, 30] are the reasons which make improvement in our R-DeCAF features. This is an important finding in which there is a linear combination among deep features which could help us to consider it in modifying networks to perform better.

Moreover, we have compared the performance of the proposed framework with the state-of-the-art studies which is summarized in Table. 4. In our method, the results of R-DeCAF features obtained by PCA algorithm have been reported. In the previous works, deep features are extracted from different pre-trained CNNs followed by SVM classifier as our case to classify BreakHis dataset. FVS is also mentioned in Table. 4 for a comprehensive analysis and comparison. As we can see, the results obtained from the proposed method have sought to increase the accuracy compared to some approaches. As a case in point, in [8] and [14], the classification of deep features which are extracted from pre-trained CNNs i.e., AlexNet, VGG-16, and VGG-19 led to lower accuracy in comparison with this study. The higher accuracy obtained by Kumar et al. [13], Gupta et al. [20], and Alinsaif et al. [16] while FVS is not comparable to our case which is almost less than

120. In [13], a global average pooling is applied to five external convolutional layers of all five blocks of VGG-16 and make a feature vector of 1472 after concatenation. This approach is a different from ours where we have just extracted features form one layer (first FC layer). This declares that looking at the features form all layers improves the result. The higher accuracy was reported in [20] where FVS is equal to 500. Although, the accuracy decreased with fewer number of features. In addition, the authors in [16] could only keep the classification accuracy unchanged by FVS equals to 1300. On the other hand, since we have only analyzed three pre-trained models as feature extractors, we could not examine our proposed concept on the mentioned works [16, 20] in which the CNNs used as a feature extractor are different. Moreover, we believe that our proposed method is able to enhance the performance of transfer learning. In another study based on fine-tuning pre-trained CNNs, the classification accuracy is reported as 80.80% for the 40× data of BreakHis dataset [5]. Reducing FVS to less than 120, our proposed method could hit the spot in comparison with previous works and classification accuracy has increased up to 4.3% simultaneously.

**Table 4.** Comparison of the classification accuracy obtained from the proposed method and previous methods.

| Existing Methods | CNN | FVS | Classification Accuracy (%) | | | | |
|---|---|---|---|---|---|---|---|
| | | | 40× | 100× | 200× | 400× | Whole |
| Bardou et al. [18] | new CNN | 2000 | 90.64 | 89.58 | 90.23 | 75.96 | - |
| Deniz et al. [8] | AlexNet + VGG-16 | 4096 + 4096 | 84.87±1.1 | 89.21±1.4 | 88.65±2.4 | 86.75±4.2 | - |
| Gupta et al. [20] | ResNet | 500 | 97.07±1.2 | 96.10±1.0 | 94.69±1.2 | 90.85±2.1 | - |
| Kumar et al. [13] | VGG-16 | 1472 | 94.11±1.8 | 95.12±1.1 | 97.01±1.1 | 93.40±1.0 | - |
| Saxena et al. [14] | AlexNet | 1526 | 84.06 | 87.54 | 89.40 | 85.16 | - |
| | VGG-16 | 3072 | 86.36 | 87.77 | 86.80 | 84.35 | - |
| | VGG-19 | 3072 | 86.64 | 88.17 | 85.84 | 81.67 | - |
| Alinsaif et al. [16] | DenseNet | 1300 | - | - | - | - | 97.96±0.6 |
| Proposed | AlexNet | 23-97 | 88.04±2.1 | 89.54±1.0 | 90.60±1.4 | 91.13±1.4 | 90.24±0.6 |
| | VGG-16 | 58-103 | 89.82±1.7 | 91.01±0.7 | 90.78±1.4 | 88.30±1.3 | 90.61±0.9 |
| | VGG-19 | 59-118 | 87.34±2.0 | 90.19±1.3 | 89.35±1.2 | 88.60±0.8 | 89.43±0.5 |

## 5. Conclusions

This study proposes R-DeCAF features for BC detection using histopathological images and compares with DeCAF features. To extract DeCAF features, three different pre-trained CNNs emerged as an unsupervised feature extractor. A feature vector from the first FC layer of CNNs with FVS of 4096 has been extracted. The results show that keeping all DeCAF features extracted from pre-trained AlexNet, VGG-16, and VGG-19 is not effective in the classification task. Thus, various dimension reduction methods on DeCAF features are evaluated to capture informative feature vectors and decrease the computational time too. Based on the analysis that we have done considering about 15% to 35% of CEV of features in the new space with FVS of less than 120 is sufficient and could significantly improve the accuracy up to 4.3% in the best case. Evaluations show that linear dimensionality reduction algorithms could represent an effective combination among deep features and lead to a higher accuracy in the classification task, however nonlinear approaches fail. This is an important finding in which there is a linear combination among deep features which could help us to consider it in modifying networks to perform better. Moreover, PCA performs better among various linear dimension reduction methods. The best-achieved result for 400× data using pre-trained AlexNet as the feature extractor is 91.13±1.4%. It should be noted that data augmentation and particular data preprocessing are not required in the proposed model that is considered a fully automatic model for cancer diagnosis. Moreover, magnification level of BreakHis dataset affect the classification accuracy as it depends on the complexity level of histopathological images. As a future work, modification in deep CNN models based on PCA algorithm to provide less features complexity and increase classification accuracy with more reliable and informative features may break this curse. Additionally, examining more other pre-trained CNN models to extract deep features and applying this proposed method for performance enhancement will be considered in the future work.

**References**

[1] T. Araújo, G. Aresta, E. Castro, J. Rouco, P. Aguiar, C. Eloy, A. Polónia, and A. Campilho, "Classification of breast cancer histology images using convolutional neural networks," *PLoS One,* vol. 12, no. 6, p. e0177544, 2017.

[2] W. H. Organization, *WHO position paper on mammography screening*. World Health Organization, 2014.

[3] P. Boyle and B. Levin, *World cancer report 2008*. IARC Press, International Agency for Research on Cancer, 2008.

[4] J. Arevalo, A. Cruz-Roa, and F. A. GONZÁLEZ O, "Histopathology image representation for automatic analysis: A state-of-the-art review," *Revista Med,* vol. 22, no. 2, pp. 79-91, 2014.

[5] S. Singh and R. Kumar, "Breast cancer detection from histopathology images with deep inception and residual blocks," *Multimedia Tools and Applications,* vol. 81, no. 4, pp. 5849-5865, 2022.

[6] F. A. Spanhol, L. S. Oliveira, P. R. Cavalin, C. Petitjean, and L. Heutte, "Deep features for breast cancer histopathological image classification," in *2017 IEEE International Conference on Systems, Man, and Cybernetics (SMC)*, 2017: IEEE, pp. 1868-1873.

[7] R. Mehra, "Breast cancer histology images classification: Training from scratch or transfer learning?," *ICT Express,* vol. 4, no. 4, pp. 247-254, 2018.

[8] E. Deniz, A. Şengür, Z. Kadiroğlu, Y. Guo, V. Bajaj, and Ü. Budak, "Transfer learning based histopathologic image classification for breast cancer detection," *Health information science and systems,* vol. 6, no. 1, pp. 1-7, 2018.

[9] G. Zhong, S. Yan, K. Huang, Y. Cai, and J. Dong, "Reducing and stretching deep convolutional activation features for accurate image classification," *Cognitive Computation,* vol. 10, no. 1, pp. 179-186, 2018.

[10] P. Filipczuk, T. Fevens, A. Krzyżak, and R. Monczak, "Computer-aided breast cancer diagnosis based on the analysis of cytological images of fine needle biopsies," *IEEE transactions on medical imaging,* vol. 32, no. 12, pp. 2169-2178, 2013.

[11] S. Sharma and R. Mehra, "Conventional machine learning and deep learning approach for multi-classification of breast cancer histopathology images—a comparative insight," *Journal of digital imaging,* vol. 33, no. 3, pp. 632-654, 2020.

[12] T. J. Alhindi, S. Kalra, K. H. Ng, A. Afrin, and H. R. Tizhoosh, "Comparing LBP, HOG and deep features for classification of histopathology images," in *2018 international joint conference on neural networks (IJCNN)*, 2018: IEEE, pp. 1-7.

[13] A. Kumar, S. K. Singh, S. Saxena, K. Lakshmanan, A. K. Sangaiah, H. Chauhan, S. Shrivastava, and R. K. Singh, "Deep feature learning for histopathological image classification of canine mammary tumors and human breast cancer," *Information Sciences,* vol. 508, pp. 405-421, 2020.

[14] S. Saxena, S. Shukla, and M. Gyanchandani, "Pre-trained convolutional neural networks as feature extractors for diagnosis of breast cancer using histopathology," *International Journal of Imaging Systems and Technology,* vol. 30, no. 3, pp. 577-591, 2020.

[15] P. Yamlome, A. D. Akwaboah, A. Marz, and M. Deo, "Convolutional neural network based breast cancer histopathology image classification," in *2020 42nd Annual International Conference of the IEEE Engineering in Medicine & Biology Society (EMBC)*, 2020: IEEE, pp. 1144-1147.

[16] S. Alinsaif and J. Lang, "Histological image classification using deep features and transfer learning," in *2020 17th Conference on Computer and Robot Vision (CRV)*, 2020: IEEE, pp. 101-108.

[17] S. Boumaraf, X. Liu, Y. Wan, Z. Zheng, C. Ferkous, X. Ma, Z. Li, and D. Bardou, "Conventional machine learning versus deep learning for magnification dependent histopathological breast cancer image classification: A comparative study with visual explanation," *Diagnostics,* vol. 11, no. 3, p. 528, 2021.

[18] D. Bardou, K. Zhang, and S. M. Ahmad, "Classification of breast cancer based on histology images using convolutional neural networks," *Ieee Access,* vol. 6, pp. 24680-24693, 2018.

[19] M. Z. Alom, C. Yakopcic, M. Nasrin, T. M. Taha, and V. K. Asari, "Breast cancer classification from histopathological images with inception recurrent residual convolutional neural network," *Journal of digital imaging,* vol. 32, no. 4, pp. 605-617, 2019.

[20] V. Gupta and A. Bhavsar, "Partially-independent framework for breast cancer histopathological image classification," in *Proceedings of the IEEE/CVF Conference on Computer Vision and Pattern Recognition Workshops*, 2019, pp. 0-0.

[21] "Breast Cancer Histopathological Database (BreakHis)." https://web.inf.ufpr.br/vri/databases/breast-cancer-histopathological-database-breakhis/ (accessed.

[22] R. F. Mansour, "Deep-learning-based automatic computer-aided diagnosis system for diabetic retinopathy," *Biomedical engineering letters,* vol. 8, no. 1, pp. 41-57, 2018.

[23] F. Anowar, S. Sadaoui, and B. Selim, "Conceptual and empirical comparison of dimensionality reduction algorithms (pca, kpca, lda, mds, svd, lle, isomap, le, ica, t-sne)," *Computer Science Review,* vol. 40, p. 100378, 2021.

[24] K. P. Murphy, *Machine learning: a probabilistic perspective*. MIT press, 2012.

[25] A. Krizhevsky, I. Sutskever, and G. E. Hinton, "Imagenet classification with deep convolutional neural networks," *Advances in neural information processing systems,* vol. 25, 2012.

[26] K. Simonyan and A. Zisserman, "Very deep convolutional networks for large-scale image recognition," *arXiv preprint arXiv:1409.1556,* 2014.

[27] L. Van Der Maaten, E. Postma, and J. Van den Herik, "Dimensionality reduction: a comparative," *J Mach Learn Res,* vol. 10, no. 66-71, p. 13, 2009.

[28] A. Tharwat, T. Gaber, A. Ibrahim, and A. E. Hassanien, "Linear discriminant analysis: A detailed tutorial," *AI communications,* vol. 30, no. 2, pp. 169-190, 2017.

[29] L. Van der Maaten and G. Hinton, "Visualizing data using t-SNE," *Journal of machine learning research,* vol. 9, no. 11, 2008.

[30] S. Karamizadeh, S. M. Abdullah, A. A. Manaf, M. Zamani, and A. Hooman, "An overview of principal component analysis," *Journal of Signal and Information Processing,* vol. 4, 2020.